%% file: 30DayReadmPaper.tex
\documentclass[english]{article}

\PassOptionsToPackage{super,comma}{natbib}
\usepackage[final]{nips_2017}

\usepackage[utf8]{inputenc} 
\usepackage[T1]{fontenc}    
\usepackage{hyperref}       
\usepackage{url}            
\usepackage{booktabs}       
\usepackage{amsfonts}       
\usepackage{nicefrac}       
\usepackage{microtype}      
\usepackage{verbatim}
\usepackage{xcolor}
\usepackage{graphicx}
\usepackage{placeins}
\usepackage[font={small}]{caption}

\usepackage{babel,blindtext}

\newif\ifcomment
\commenttrue  

\title{Predicting readmission risk from doctors' notes}

\author{
  Erin ~Craig\\
  Florence A. Rothman Institute\\
  1803 Glengary Street\\
  Sarasota, FL 34231\\
  \texttt{erin.craig@farinstitute.org} \\
  \And
  Carlos ~Arias\\
  Florence A. Rothman Institute\\
  1803 Glengary Street\\
  Sarasota, FL 34231\\
  \texttt{carlos.arias@farinstitute.org} \\
  \And
  David ~Gillman \\
  New College of Florida\\
  5800 Bay Shore Road\\
  Sarasota FL 34243 \\
  \texttt{dgillman@ncf.edu} \\ 
}

\begin{document}


\maketitle

\input{abstract}

\section{Introduction}
\input{intro}

\section{Methods and Data}
\input{methods}

\section{Results}
\input{results}

\FloatBarrier
\section{Discussion}
\input{discussion}

\section{Acknowledgments}
\input{acknowledgments}



\bibliographystyle{vancouver}
\bibliography{readmission}

\FloatBarrier
\section{Appendix}
\input{figures}

\FloatBarrier

\medskip

\small

\end{document}

%% file: abstract.tex
\begin{abstract}
  
We develop a model using deep learning techniques and natural language processing on unstructured text from medical records to predict hospital-wide $30$-day unplanned readmission, with c-statistic $.70$. Our model is constructed to allow physicians to interpret the significant features for prediction.

\end{abstract}

%% file: intro.tex
Hospital readmission is bad for the health of patients and costly to the healthcare system. \cite{Jencks2009,Bueno2010} Cases of unavoidable readmission exist, but the variation of readmission rates across hospitals suggests that some cases are predictable and avoidable. \cite{Zhang2008} In 2012 the Affordable Care Act enacted the Hospital Readmissions Reduction Program as an incentive for hospitals to reduce readmissions for some conditions. \cite{CMSReadmissions2016}

In this study we consider hospital-wide unplanned readmission for acute care within $30$ days of discharge. The Centers for Medicare and Medicaid Services (CMS) define an unplanned readmission as a readmission for unscheduled acute care that is not for an organ transplant, chemotherapy, or radiation, or a potentially planned visit that includes acute or complication of care. \cite{CMSMethodology2017} The definition applies to hospital-wide ("all-cause") visits. Admissions within $24$ hours for the same condition are not considered readmissions by the CMS.  

Research on unplanned readmission has sought to identify features of patients that put them at risk. The systematic reviews of Kansagara et al. (2011) and Zhou et al. (2016) show increased interest in predictive models, citing $14$ models from 2011 to 2015 that predict $30$-day hospital-wide unplanned readmission with c-statistics ranging from $.55$ to $.79$. \cite{Kansagara2011, Zhou2016}  

These models are valuable for two purposes: for use in clinical tools that flag at-risk patients for intervention during their hospital stay to reduce their risk of readmission, and for retrospective analysis of the causes of readmission. Models for early detection use features such as demographics, admission diagnosis and acuity, and prior hospital visits. Models for retrospective analysis use these features as well as administrative data available at the time of billing, such as comorbidities, length of stay, procedures, and prescriptions. The LACE score (length of stay, acuity of admission, comorbidities, prior emergency visits) is used by many models. \cite{vanWalraven2010}

Increasingly, models also use clinical information, such as laboratory tests, or patient surveys. \cite{Coleman2004,Escobar2015,Richmond2013} 
Administrative data and clinical information are available as structured entries in the electronic medical record (EMR), and computational methods render that data available sufficiently early for clinical use. \cite{Escobar2015} However, while structured data contain features that are well-established predictors of readmission, they offer an incomplete picture of the patient. An example is the presence of comorbidities. Patients may have preexisting conditions for which they received no treatment during the index hospitalization. Diabetes and dementia are common examples.

Doctors' notes hold information about a patient that cannot always be found in administrative data, laboratory reports, or nursing evaluations. Physicians often describe the patient history and ongoing conditions, the treatments and procedures attempted during the index hospitalization together with their success or failure, and discharge instructions that indicate whether the patient is able to resume normal activity. Finally, physicians occasionally write a direct opinion on the patient's prognosis.

We are interested in extracting information from the unstructured text in medical records using deep learning models and natural language processing (NLP). Often these models have strong predictive power but their predictions resist interpretation. \cite{Breiman2001} A model is more useful as a clinical tool if the physician understands the features underlying its predictions. In this work we demonstrate that it is feasible to extract meaningful signal from unstructured text using a deep learning model that lends itself to interpretation by physicians. The information derived from our model is informative for detecting and intervening on behalf of patients at risk of unplanned readmission hospital-wide.  

We trained a convolutional neural network (CNN) to predict hospital-wide $30$-day unplanned readmission using the text of doctors' discharge notes. We used simple preprocessing to segment the text into sections and lists. We also used standard techniques to convert words of text to vectors for input to the CNN. To overcome the black box nature of neural networks, we structured our model to allow visualization and explanation of its predictions. The c-statistic for our model was $.7$ on test data, as compared to $.65$ for a logistic regression on the features used in the LACE score.

NLP has been applied in the medical domain for some time. \cite{Agarwal2017, Wallace2017, HARVEST2015, Dligach2014, Rumshisky2016} Research has applied standard machine learning techniques and NLP on unstructured text to predict readmissions for certain conditions \cite{Watson2011,Wasafy2015} and hospital-wide \cite{Greenwald2017}. Recent work has applied deep learning techniques and NLP to characterize patient phenotype for the purpose of individualized patient care. \cite{Gehrmann2017} Other recent work has applied deep learning techniques and NLP to predict readmissions for diabetes patients. \cite{Duggal2016}

%% file: methods.tex
Our goal is to predict $30$-day, hospital-wide, unplanned hospital readmissions at Sarasota Memorial Hospital. We classify a readmission as planned or unplanned using the planned readmission algorithm \cite{CMSMethodology2017} developed by CMS. We note that this hospital sets itself apart as one of the $4$\% of U.S. hospitals with a readmission rate lower than the national readmission rate ($13.6$\% vs $15.3$\%). \cite{CMSReadmission2016} Our patient population has ages ranging from $29$ to $108$ (median: $71$, mean: $69$), and $91\%$ of patients self-identify as White or Caucasian. 
   
Our model uses physician's discharge notes from $141,226$ inpatient visits during the years 2004 to 2014. We omitted visits where the patient was discharged to hospice or passed away within $30$ days of discharge. We also omitted visits where the discharge note had fewer than $20$ words. This study was approved by the hospital's Institutional Review Board.

To develop and test our model, we split the data into three groups: a training set ($113,077$ visits), a validation set ($12,566$ rows) and a test set ($15,583$ visits). 

\subsection{Data preparation}
        We preprocessed the discharge notes before passing them into our neural network. We removed names, numbers, dates, and punctuation to minimize overfitting and bias. To give all notes the same structure, we labeled and reordered sections (e.g. allergies, prognosis, discharge condition). This was done by manually inspecting notes for consistent section headers, and using pattern matching to extract the sections. Finally, we imposed a length of $700$ words on each discharge note by removing words from the end of longer notes, and right-padding shorter notes with the string \texttt{PADDING}.

\subsection{Model}
\begin{figure}[!htbp]
\begin{center}
\includegraphics[width = 0.7\textwidth]{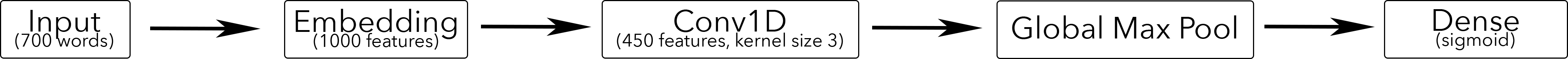}
\caption{Model architecture}
\label{fig:modelarchitecture}
\end{center}
\end{figure}

        After trying several architectures, we found that a one-dimensional convolutional neural network achieved the highest c-statistic. Our model consisted of a word embedding layer (initialized with a pre-trained word embedding from word2vec skip-gram with negative sampling), a convolutional layer, a max pooling layer and a dense layer (figure \ref{fig:modelarchitecture}). Training was done using the Keras framework\cite{chollet2015keras} running TensorFlow\cite{tensorflow2015-whitepaper} as the backend, with RMSprop\cite{rmsprop} as the optimizer. For full details and model hyperparameters, we refer to our code\footnote{\url{https://github.com/farinstitute/ReadmissionRiskDoctorNotes}}. 

\subsection{Interpretability}
To retain model interpretability, we devised a shallow model. The output layer acts on the max pooling layer, and each node in the max pooling layer corresponds directly to a single trigram from the discharge note. As a result of this structure, it is possible to identify the phrases in the text that most influenced the model. 

\begin{figure}[!htbp]
\begin{center}
\includegraphics[width = 0.8\textwidth]{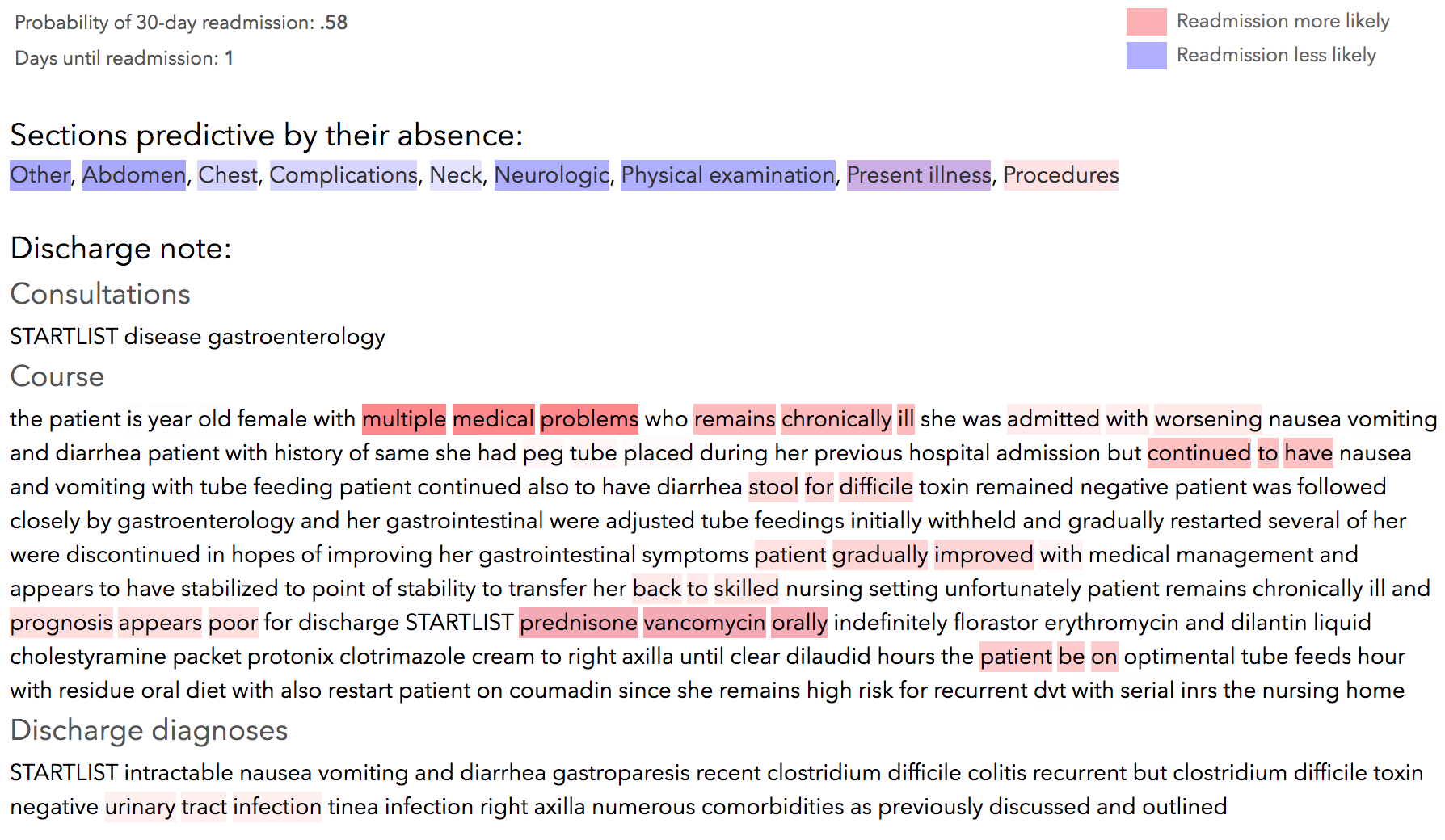}
\caption{Predicted readmission, true positive.}
\label{fig:example1}
\end{center}
\end{figure}

For more example discharge notes, refer to figures \ref{fig:example2} through \ref{fig:example7}. These examples represent the extremes in our data: they include patients who were readmitted on the same day as well as patients who were never readmitted, and they show model predictions ranging from $.04$ to $.68$.

%% file: results.tex
On test data, our model achieved a c-statistic of $.70$; see Table \ref{baselines} for comparison with other models.

\begin{table}[!htbp]
\caption{Model Performance on Test Data}\label{baselines}

\begin{tabular}{l|l|r} 
      \textbf{Model} & \textbf{Data} & \textbf{C-statistic}\\
      \hline
      \textbf{1-D convolutional neural network} & \textbf{Discharge note} & $\textbf{.70}$\\
      1-D convolutional neural network & Discharge note, LACE features & $.70$\\
      Random forest & Discharge note (TF-IDF matrix) & $.67$\\
      2-layer feed forward neural network & LACE features, LACE score & $.66$\\
      Logistic regression & LACE features, LACE score & $.66$\\
    \end{tabular}
\end{table}

%% file: discussion.tex
While our model does not improve upon the predictive performance of all of the published models, it compares favorably with most. More importantly, it demonstrates that unstructured text in the electronic medical record contains a meaningful signal that is accessible through a neural network model that is shallow enough to remain interpretable.

\begin{figure}[!htb]
\minipage{0.32\textwidth}
  \includegraphics[width=\linewidth]{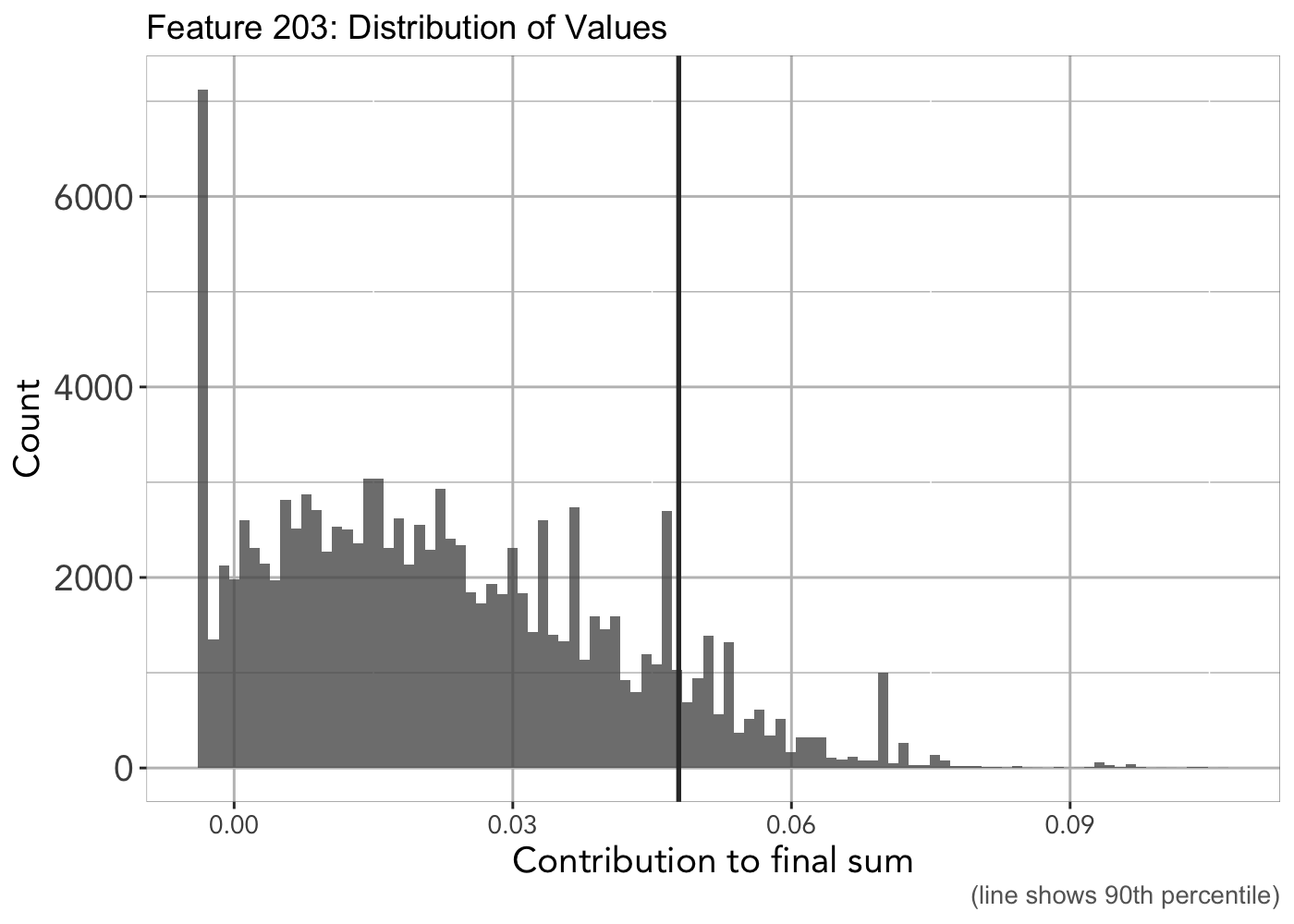}
\endminipage\hfill
\minipage{0.32\textwidth}
  \includegraphics[width=\linewidth]{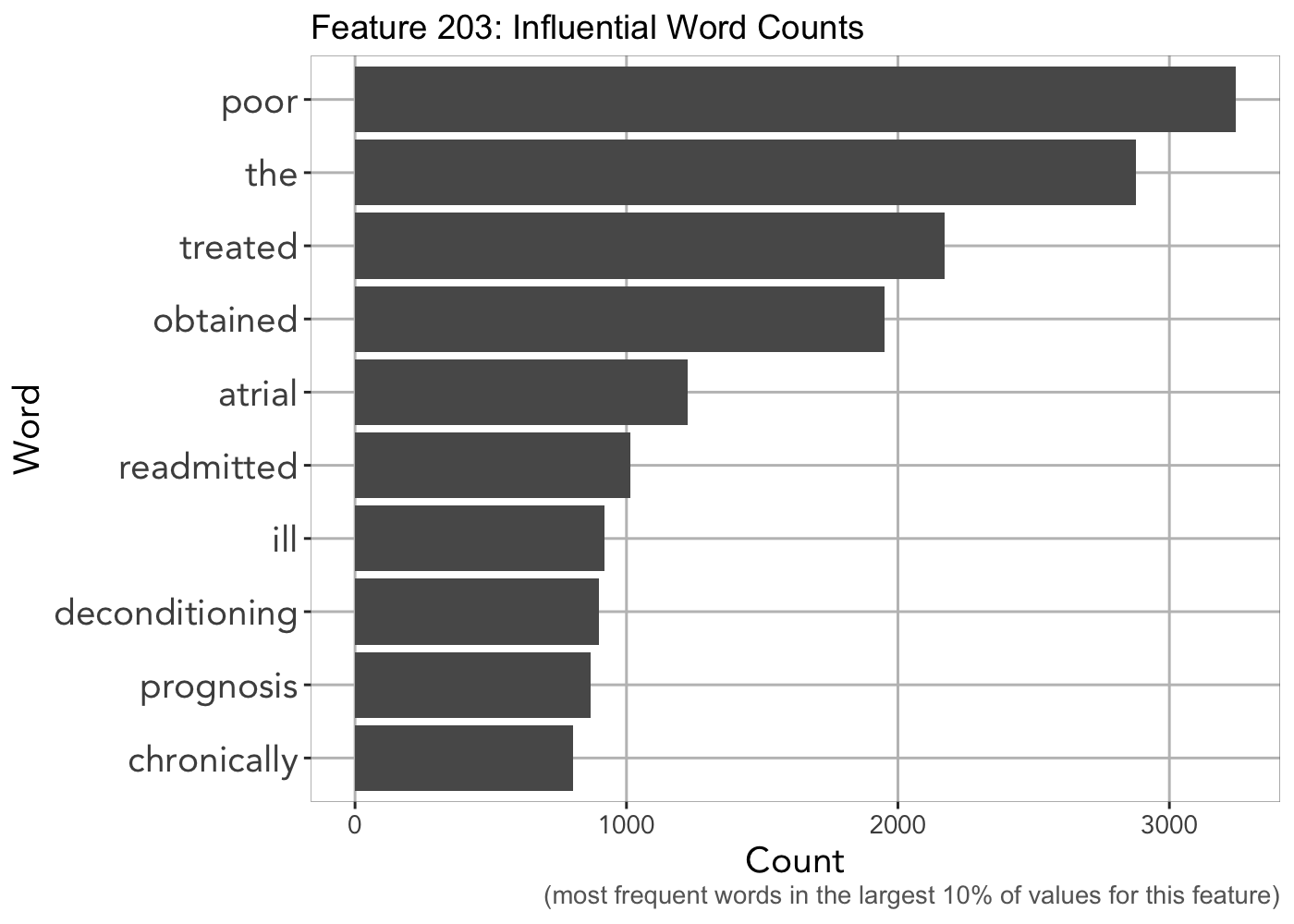}
\endminipage\hfill
\minipage{0.32\textwidth}%
  \includegraphics[width=\linewidth]{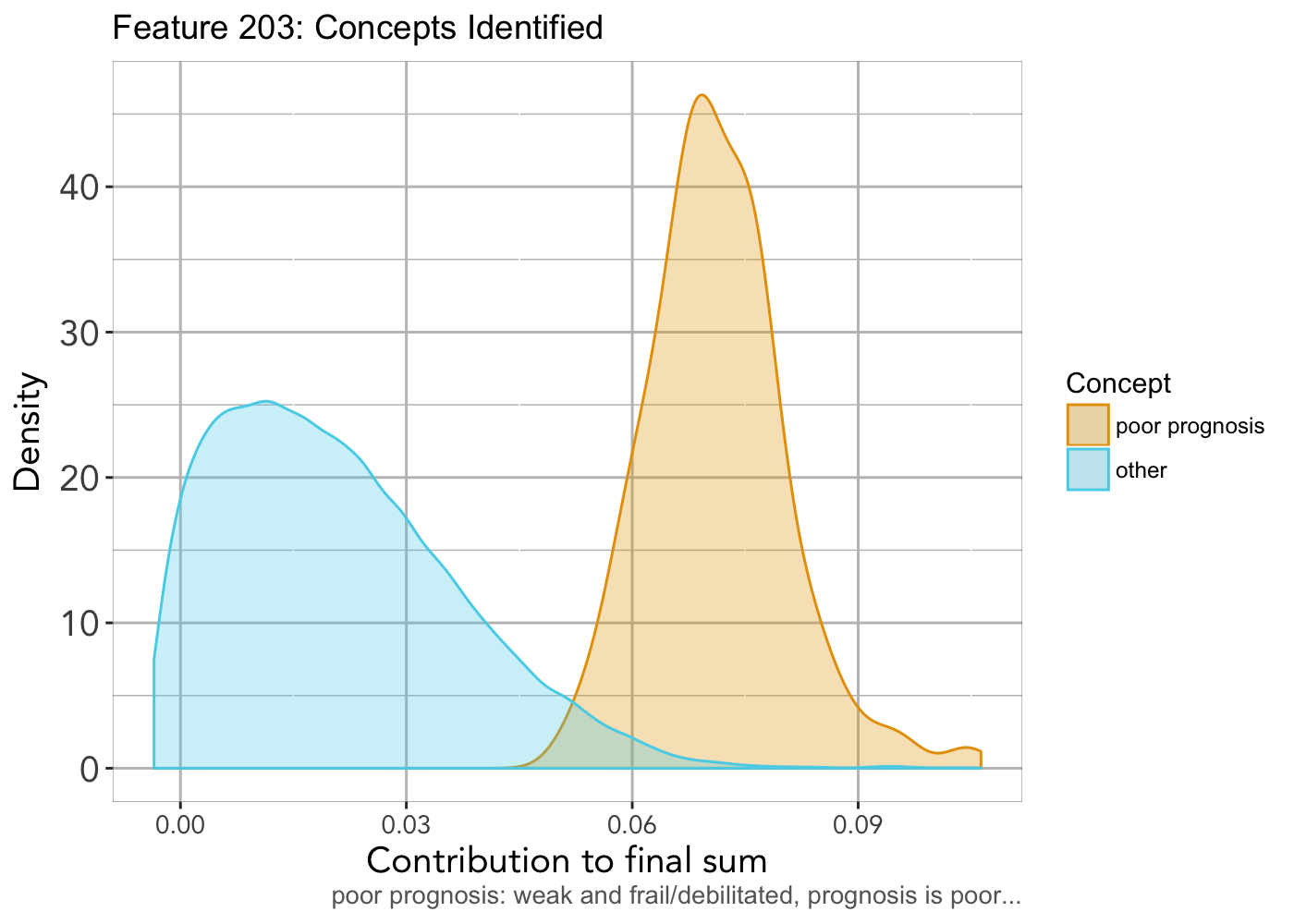}
\endminipage
\caption{Learned feature identifying clinician opinion of a poor prognosis.}\label{fig:opinions}
\end{figure}

We passed our training data through the model and recorded the values at each node in the max pooling layer, multiplied by the corresponding weight for the following dense layer. We consider these distributions to be the "contribution" of each node to the final sum and sigmoid that computes the probability of readmission. We sampled a subset of nodes for further study: we looked at those with the largest absolute values and the largest standard deviations. Then, given a single node, we identified the words corresponding to its most extreme contributions; these are the words (and topics) that "light up" this node.  From this cursory study, we found that our model's learned features identified clinician opinions (figure \ref{fig:opinions}), procedures performed (figure \ref{fig:procedures}), categories of drugs (figure \ref{fig:steroids}) and ongoing conditions (figure \ref{fig:ongoing}). Some of these features are redundant with other medical data; for example, a procedure performed will appear in a patient's clinical orders and billing data. However, the model also identified features unique to clinical text, including ongoing conditions and clinical opinions.

We note that our study has limitations. Our data originates from a single hospital, so patients who were readmitted within $30$ days to another hospital are incorrectly labeled. Another limitation of our data is that $40\%$ of discharge notes are written after discharge, thereby impeding our model's ability to provide timely decision support for all patients. 

We envision further research that could improve our model's utility. Our model's predictive power might improve through the incorporation of quantitative data from the electronic medical record, or through restriction to patients with particular conditions and procedures. It may also benefit from the application of further NLP techniques to make the text's signal more readily available to the neural network.  

This model also offers opportunity for new administrative and clinical insights through a thorough study of its learned features. 



%% file: acknowledgments.tex
We would like to thank our colleagues at the Florence A. Rothman Institute: Duncan Finlay, Steven Rothman and Robert Smith. This research was made possible by their medical expertise and support.

%% file: figures.tex



\offinterlineskip

\begin{figure}[!htbp]
\begin{center}
\includegraphics[width = 0.8\textwidth]{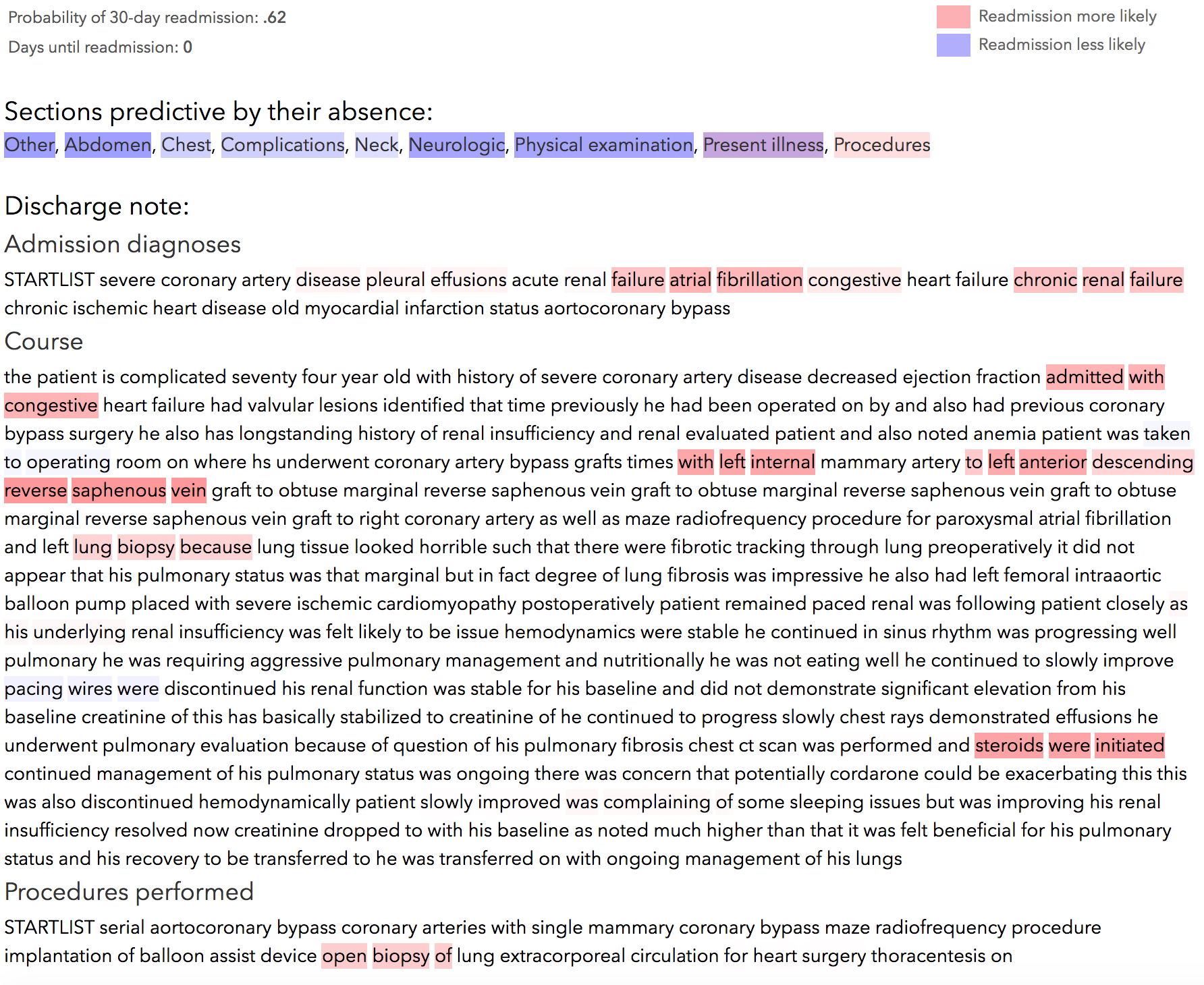}
\caption{Predicted readmission, true positive.}
\label{fig:example2}
\end{center}
\end{figure}

\begin{figure}[!htbp]
\begin{center}
\includegraphics[width = 0.8\textwidth]{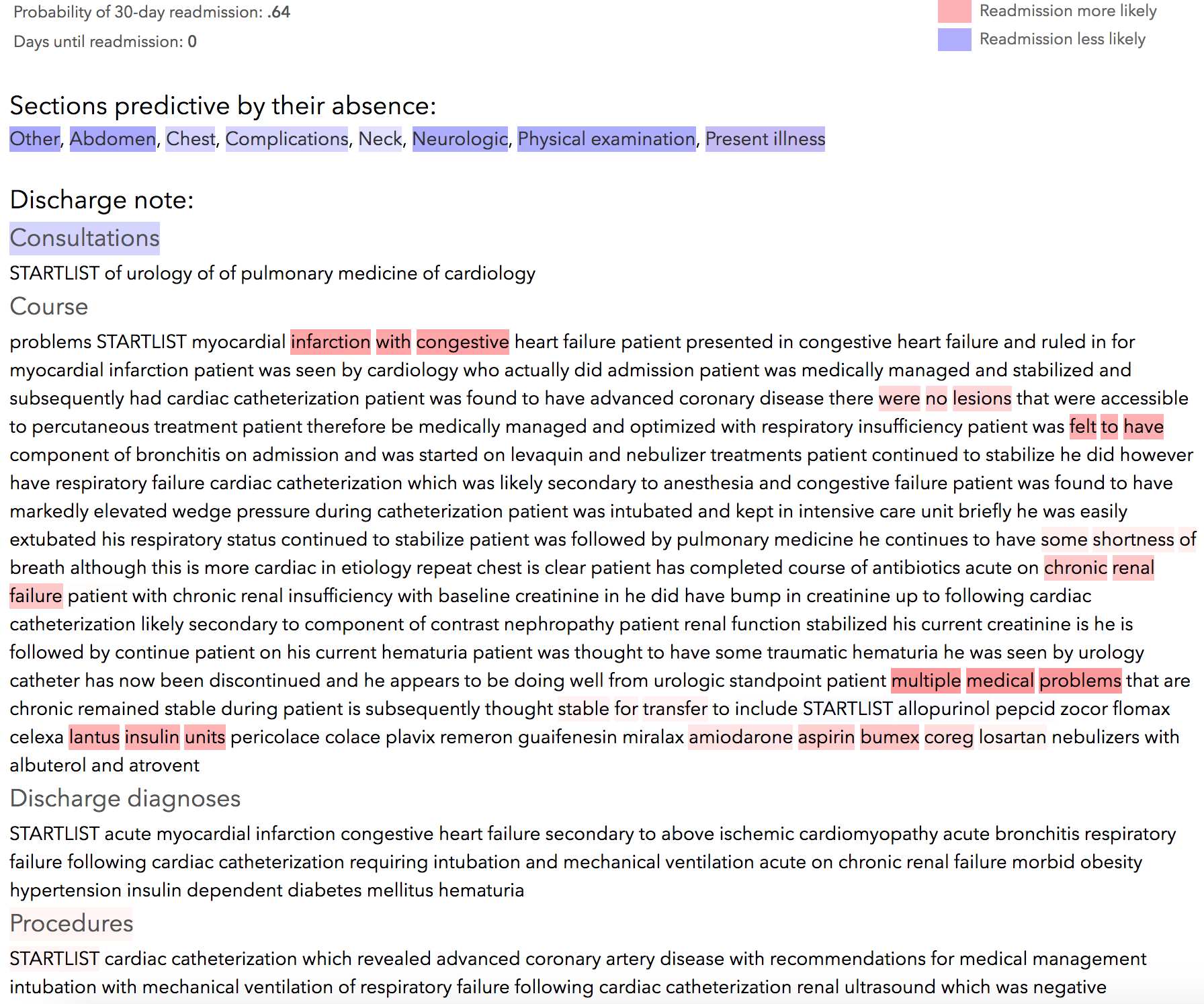}
\caption{Predicted readmission, true positive.}
\label{fig:example3}
\end{center}
\end{figure}

\begin{figure}[!htbp]
\begin{center}
\includegraphics[width = 0.8\textwidth]{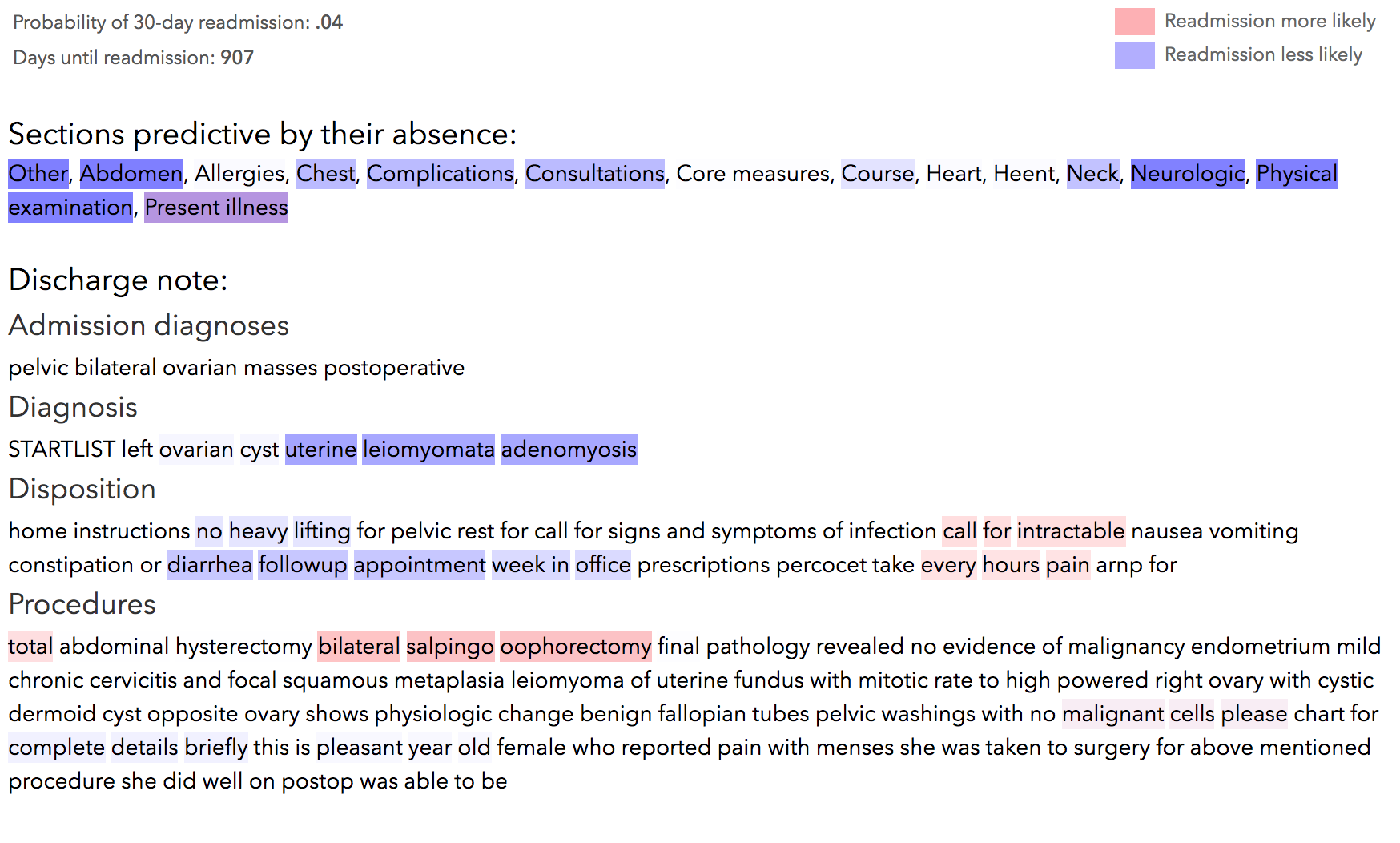}
\caption{Predicted no readmission, true negative.}
\label{fig:example4}
\end{center}
\end{figure}



\begin{figure}[!htbp]
\begin{center}
\includegraphics[width = 0.8\textwidth]{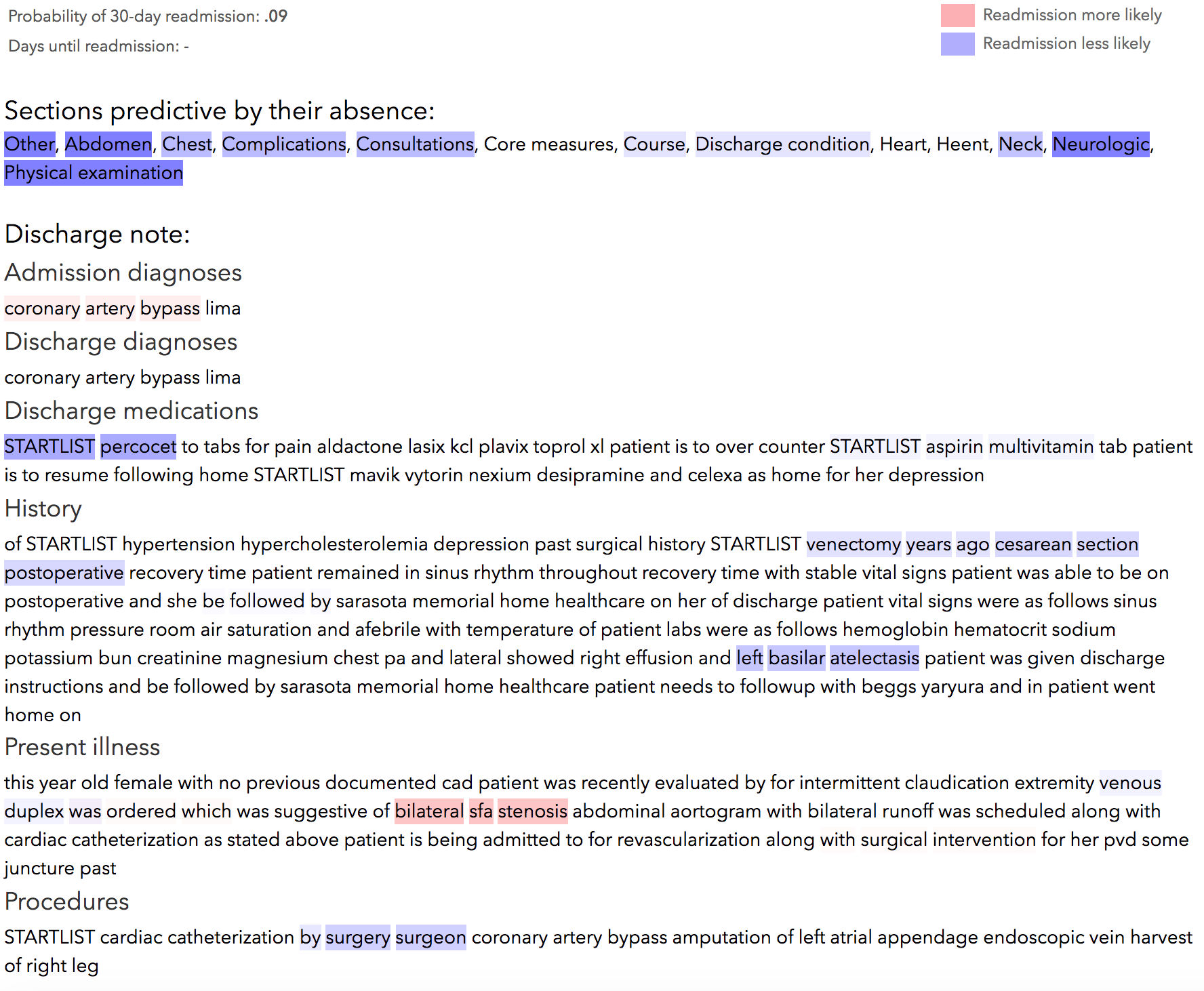}
\caption{Predicted no readmission, true negative.}
\label{fig:example5}
\end{center}
\end{figure}

\begin{figure}[!htbp]
\begin{center}
\includegraphics[width = 0.8\textwidth]{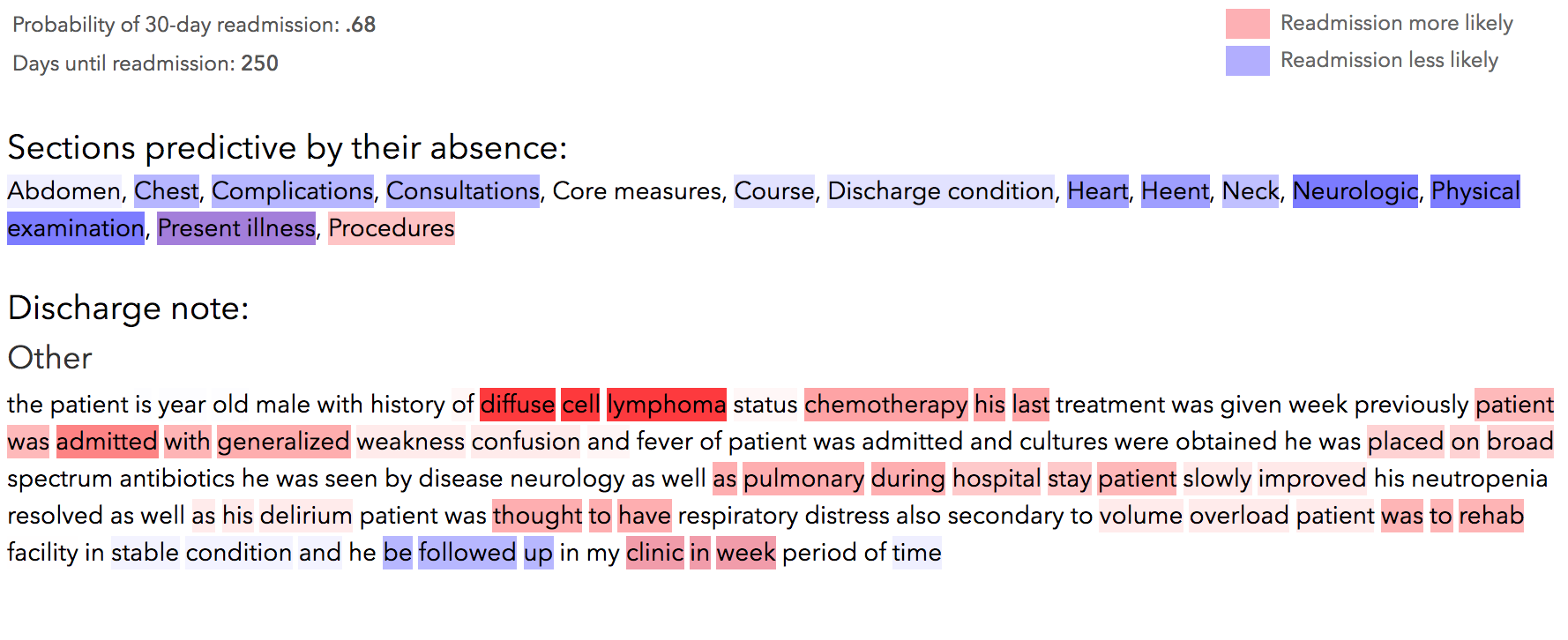}
\caption{Predicted readmission, false positive.}
\label{fig:example6}
\end{center}
\end{figure}

\begin{figure}[!htbp]
\begin{center}
\includegraphics[width = 0.8\textwidth]{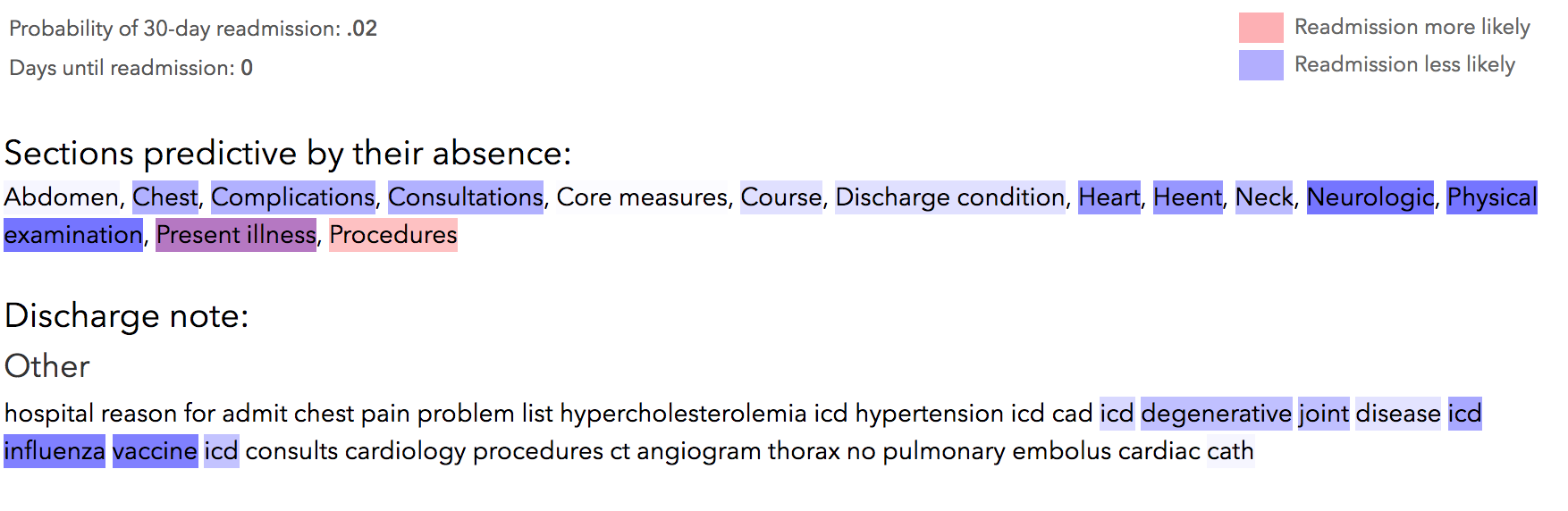}
\caption{Predicted no readmission, false negative.}
\label{fig:example7}
\end{center}
\end{figure}


\begin{figure}[!htb]
\minipage{0.32\textwidth}
  \includegraphics[width=\linewidth]{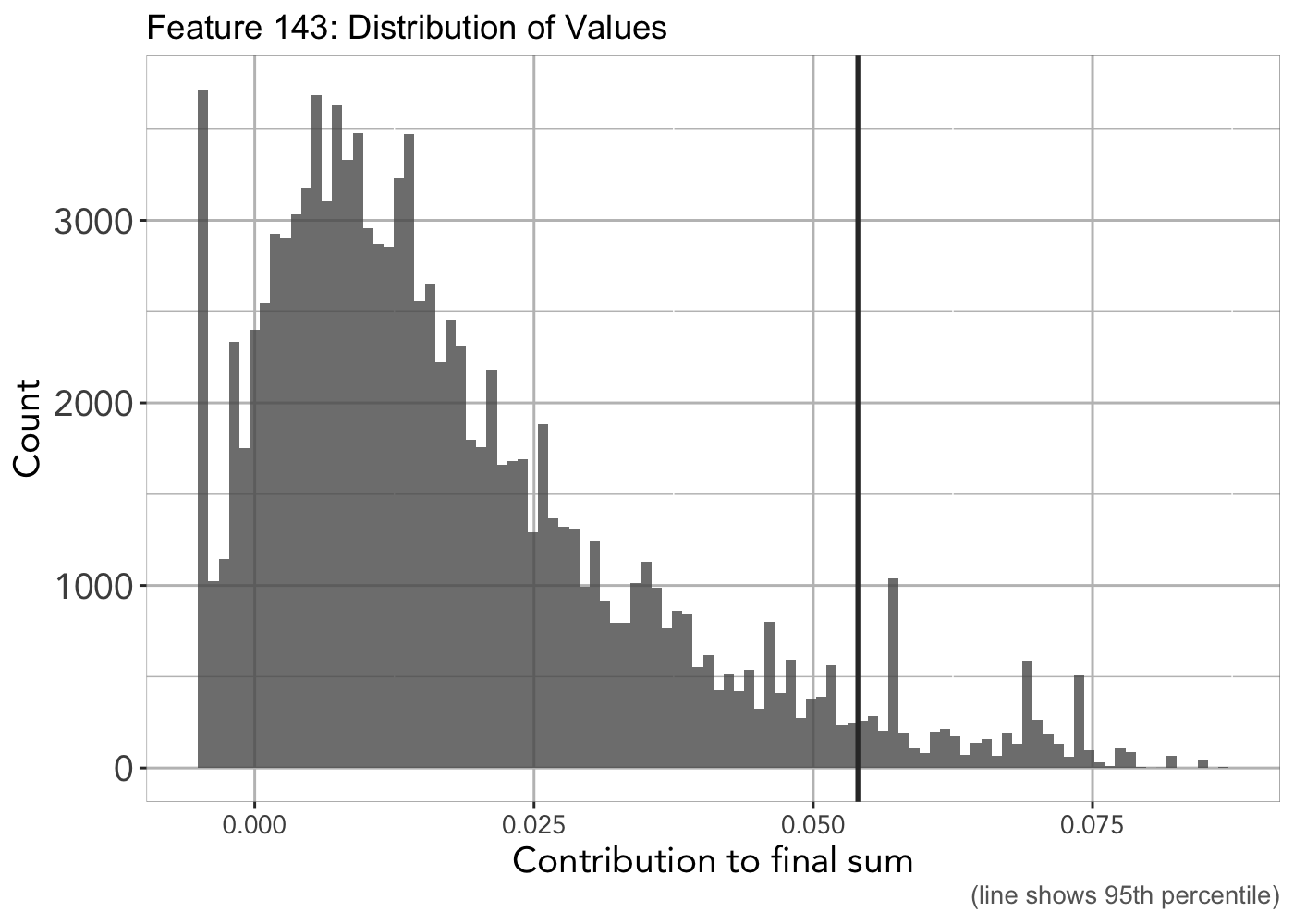}
\endminipage\hfill
\minipage{0.32\textwidth}
  \includegraphics[width=\linewidth]{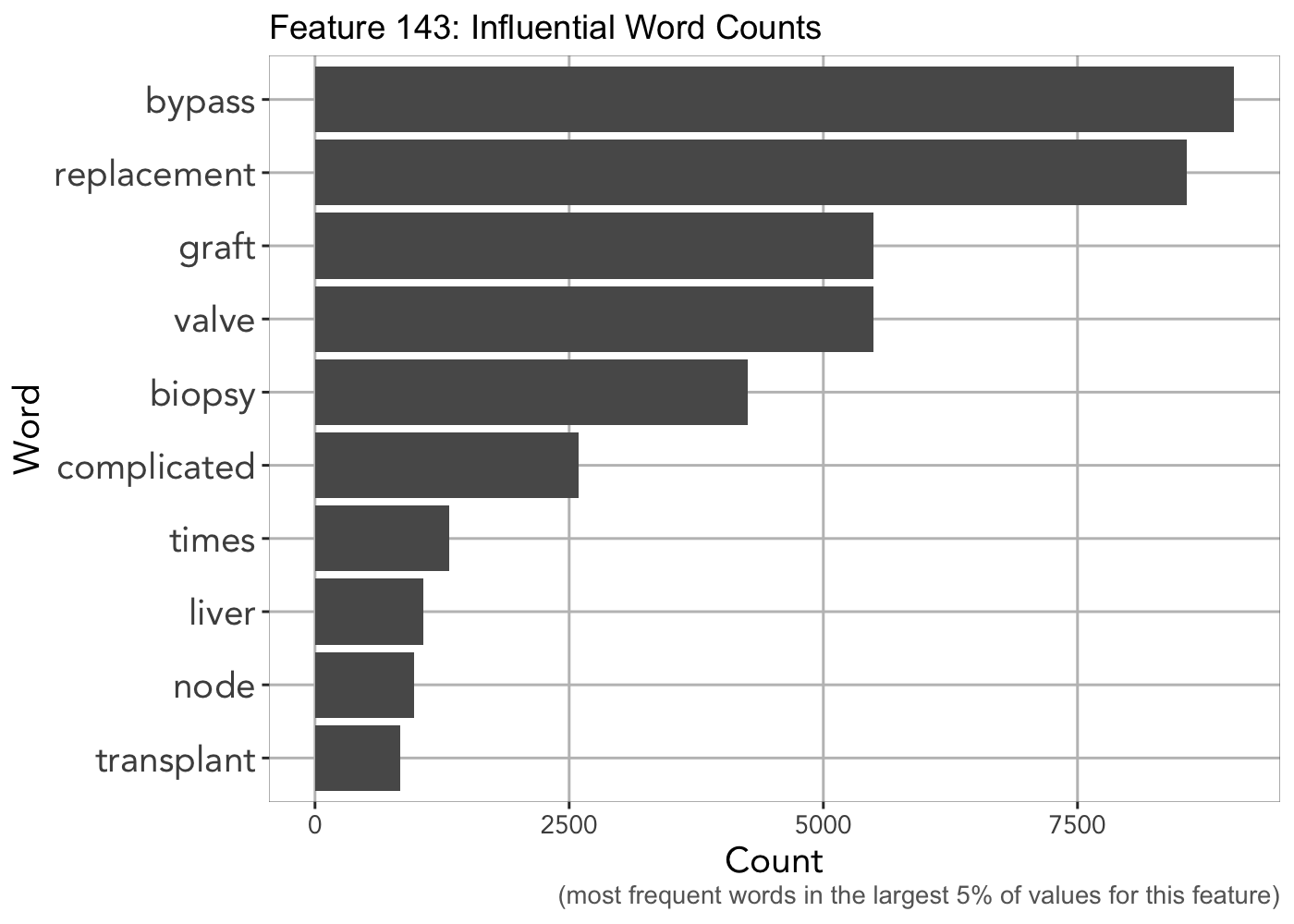}
\endminipage\hfill
\minipage{0.32\textwidth}%
  \includegraphics[width=\linewidth]{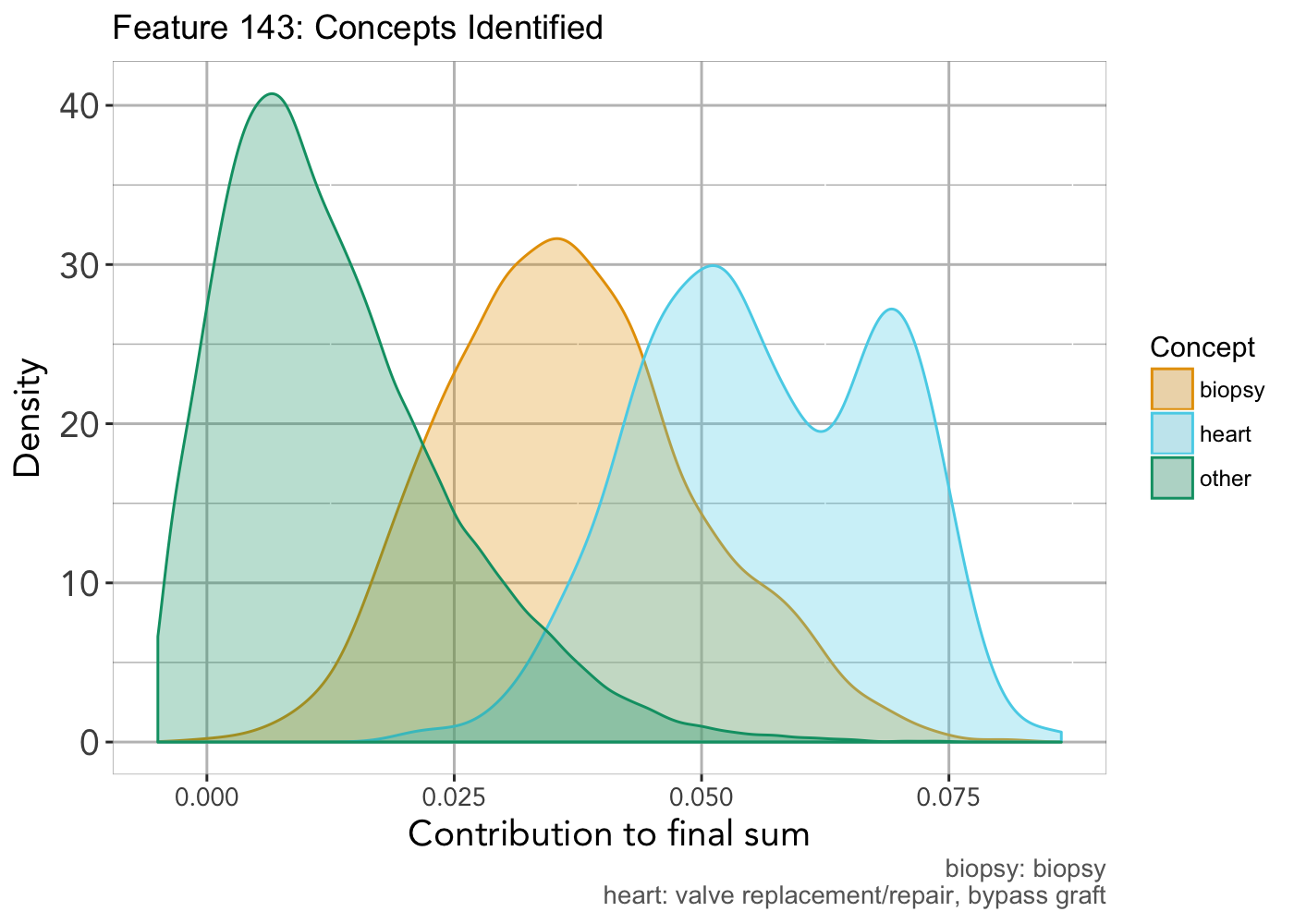}
\endminipage
\caption{Learned feature identifying a biopsy or heart surgery.}\label{fig:procedures}
\end{figure}

\begin{figure}[!htb]
\minipage{0.32\textwidth}
  \includegraphics[width=\linewidth]{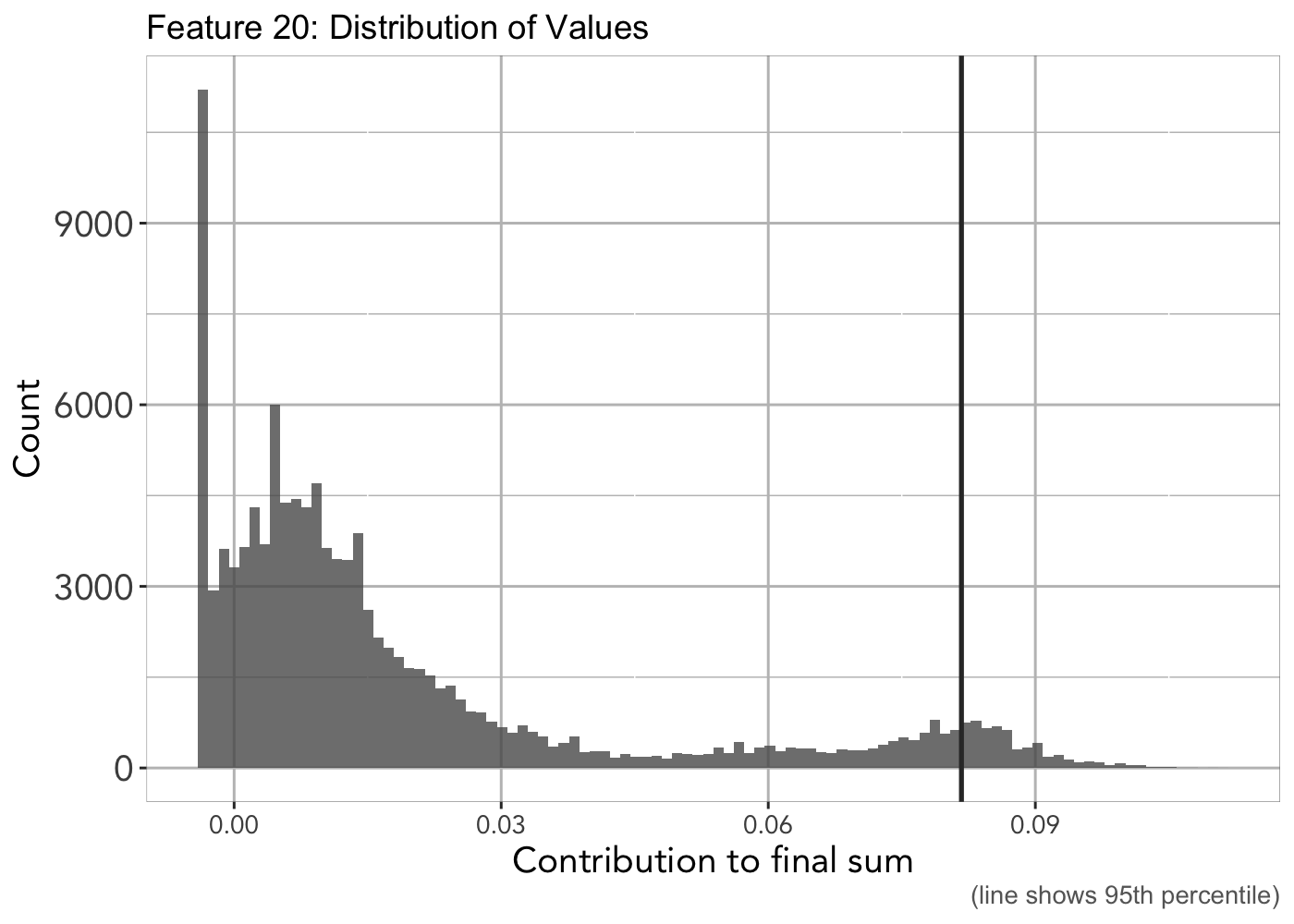}
\endminipage\hfill
\minipage{0.32\textwidth}
  \includegraphics[width=\linewidth]{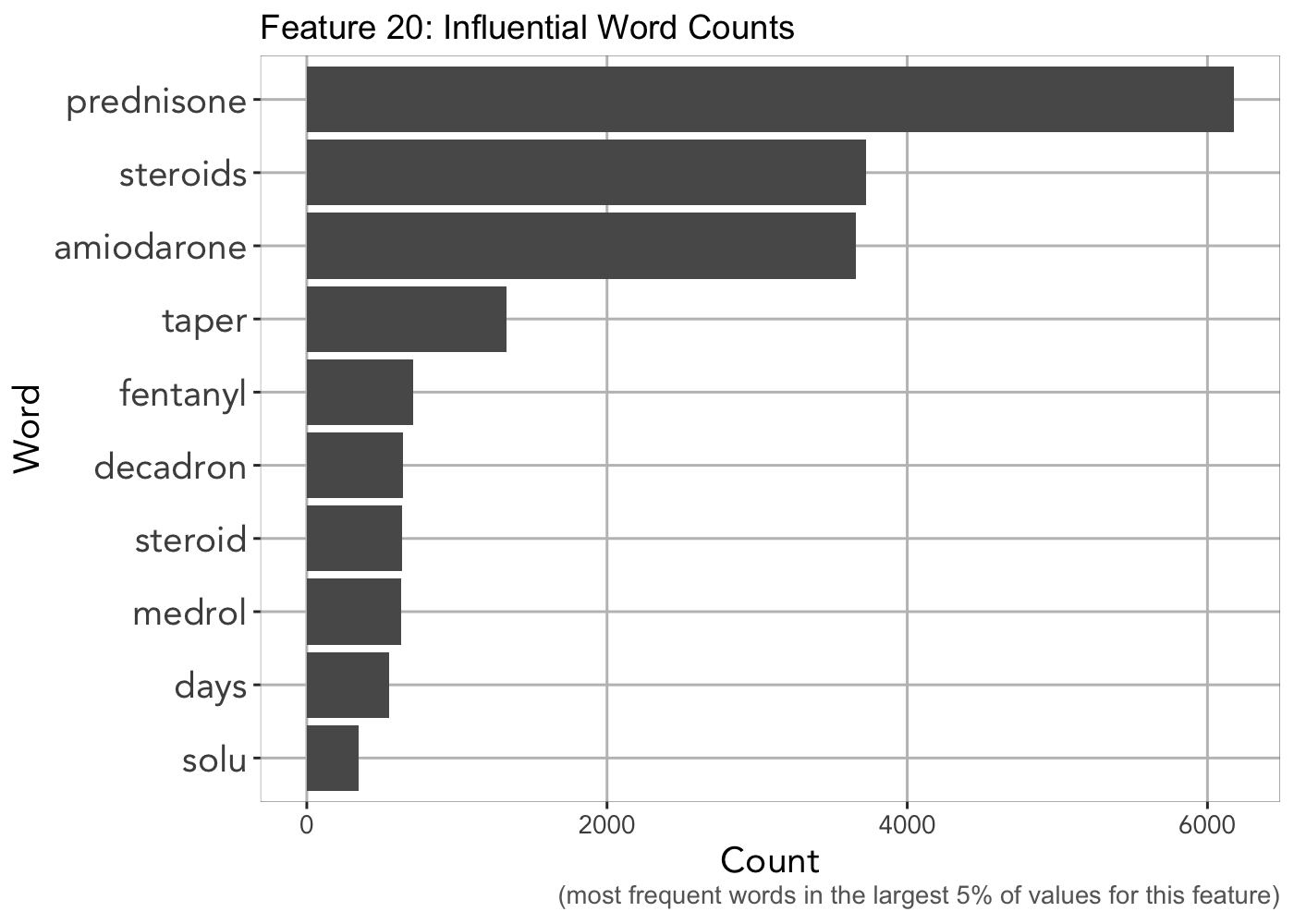}
\endminipage\hfill
\minipage{0.32\textwidth}%
  \includegraphics[width=\linewidth]{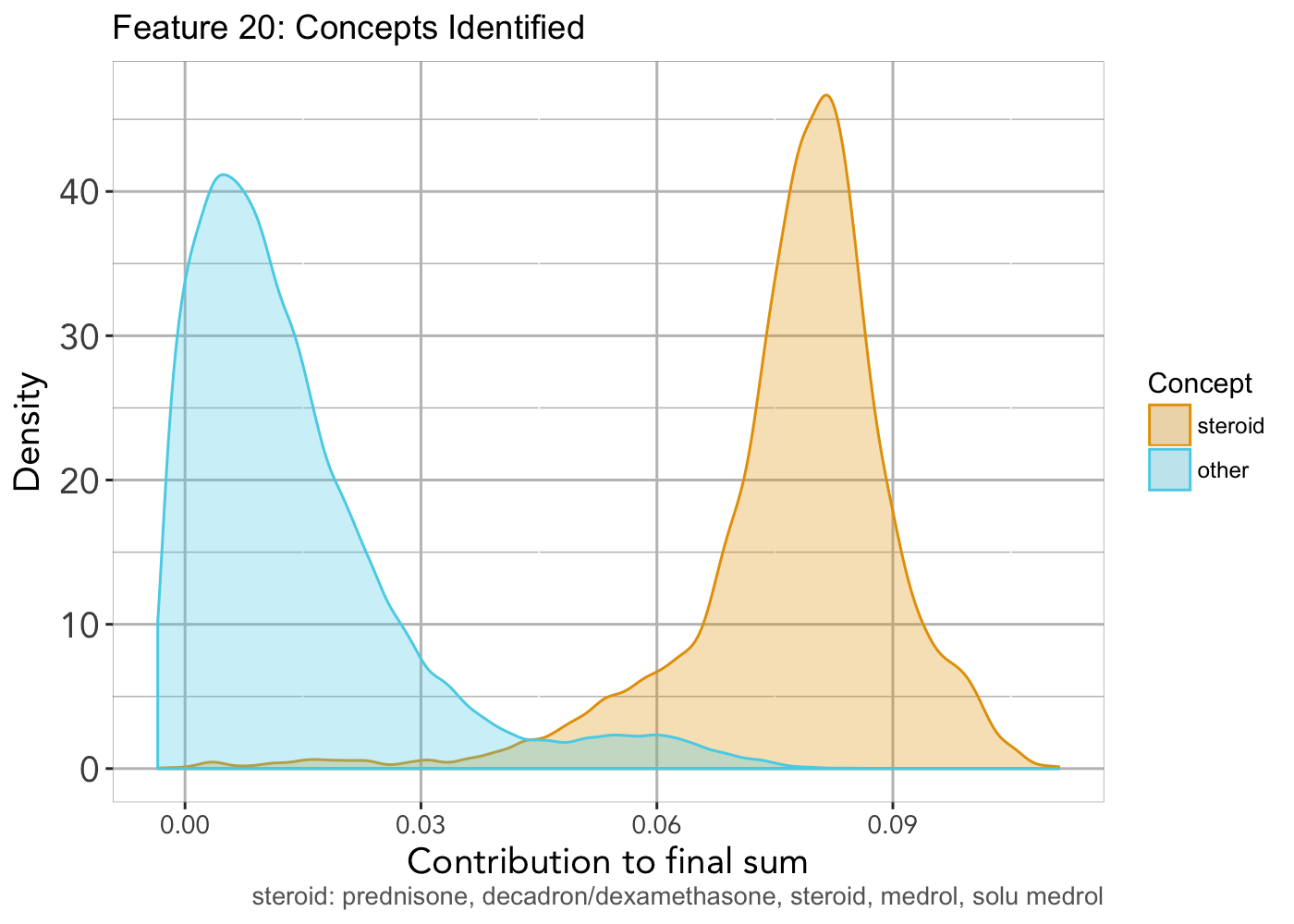}
\endminipage
\caption{Learned feature identifying presence of steroids in a patient's chart.}\label{fig:steroids}
\end{figure}

\begin{figure}[!htb]
\minipage{0.32\textwidth}
  \includegraphics[width=\linewidth]{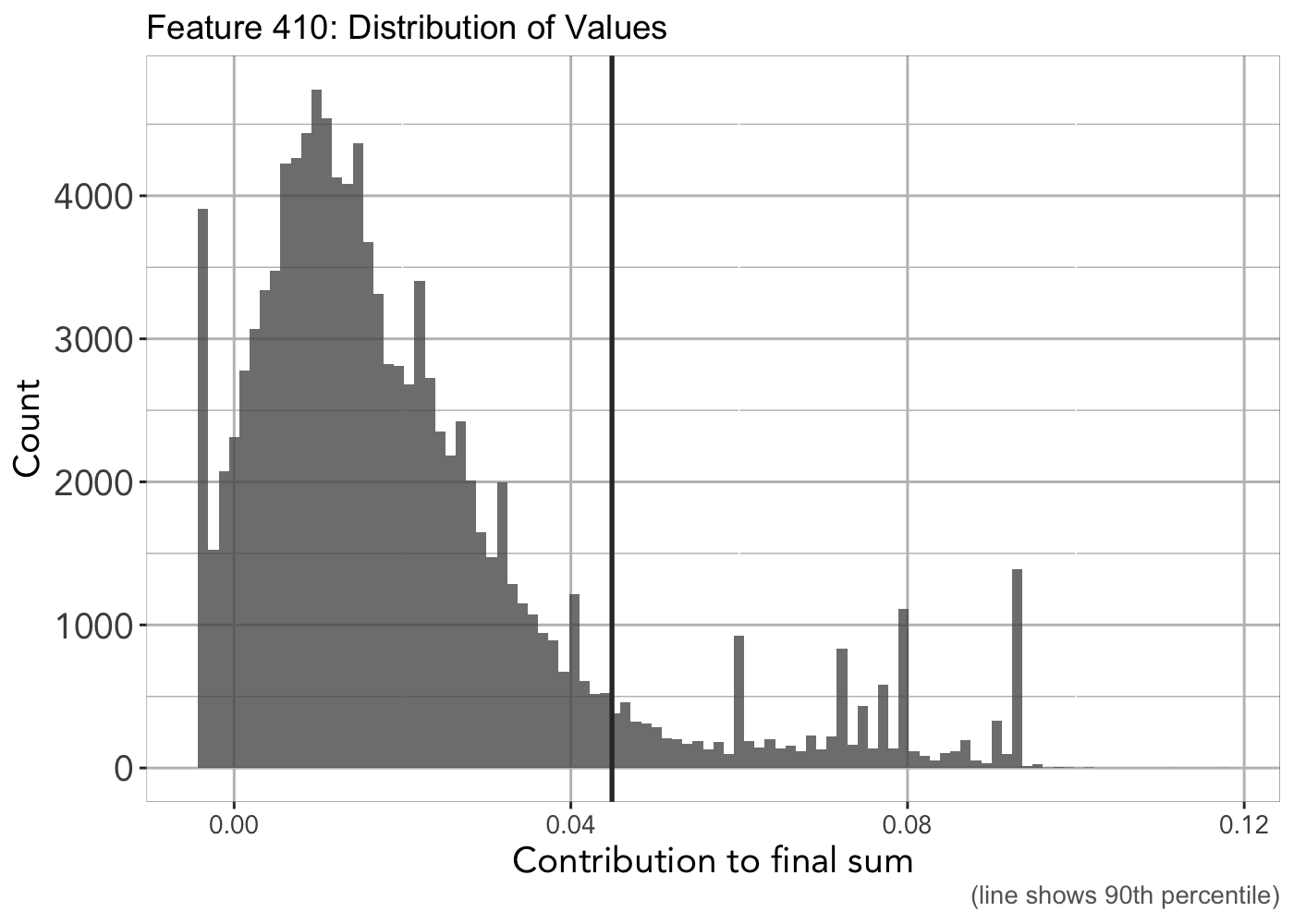}
\endminipage\hfill
\minipage{0.32\textwidth}
  \includegraphics[width=\linewidth]{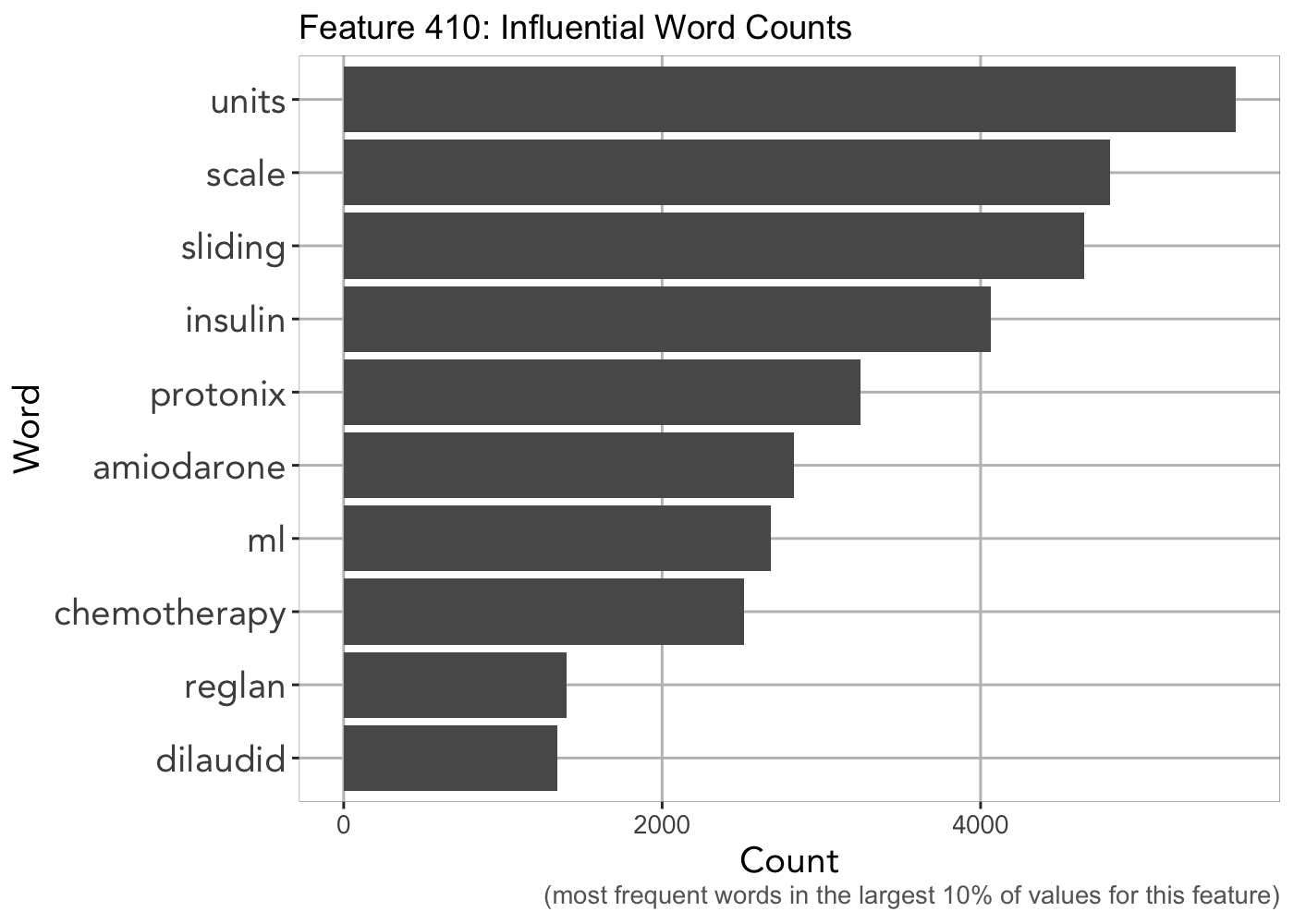}
\endminipage\hfill
\minipage{0.32\textwidth}%
  \includegraphics[width=\linewidth]{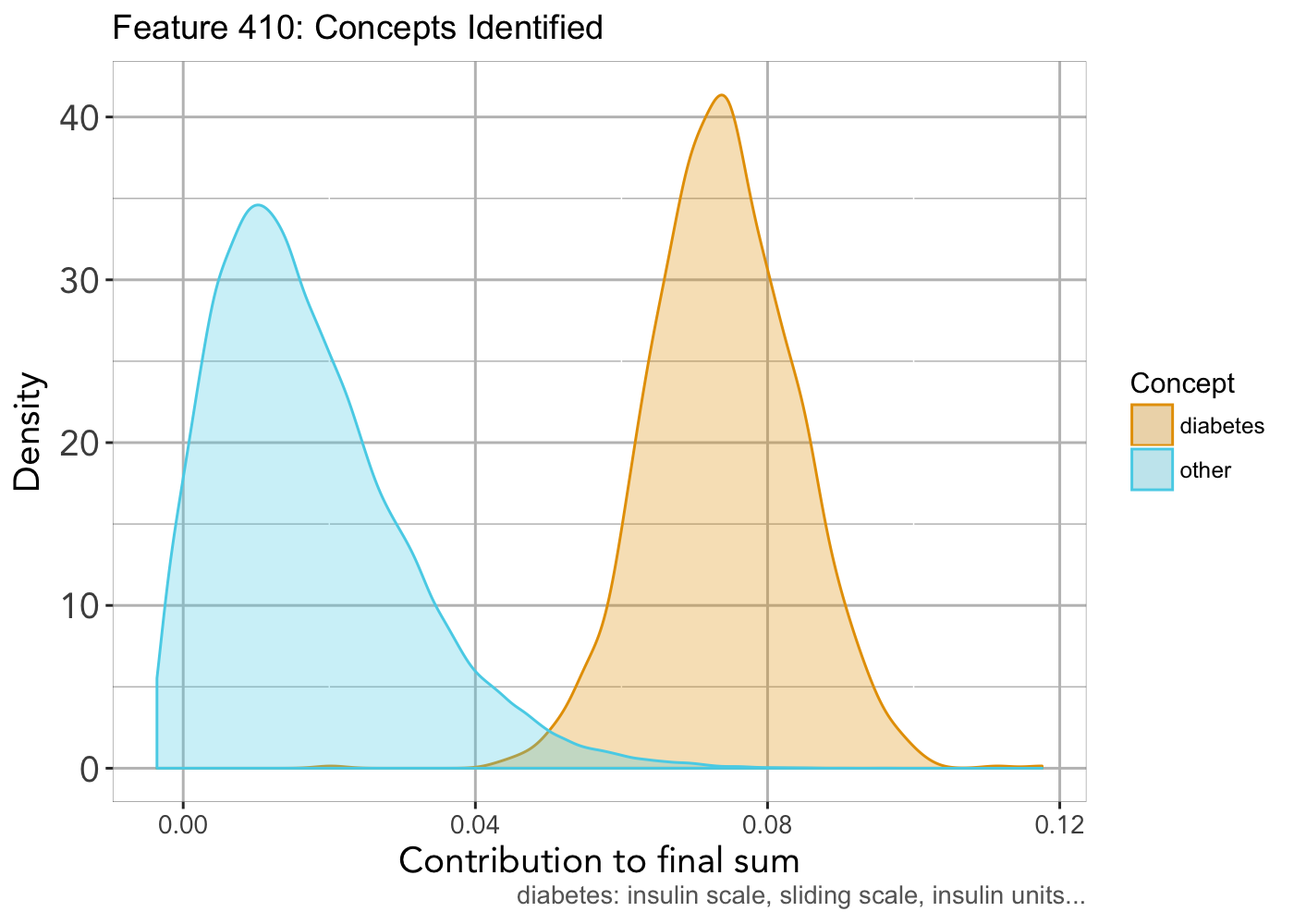}
\endminipage
\caption{Learned feature identifying ongoing diabetes.}\label{fig:ongoing}
\end{figure}